\documentclass[11pt]{article}
\usepackage{acl2016}
\usepackage{times}
\usepackage{latexsym}
\usepackage{url}
\usepackage{booktabs}
\usepackage{graphicx}
\usepackage{color}
\usepackage{amsmath}
\usepackage{todonotes}
\usepackage{multirow}
\usepackage{makecell}
\usepackage{float}
\usepackage{array}
\usepackage{microtype}

\usepackage{natbib}
\setcitestyle{comma, numbers,sort&compress, super}

\usepackage{url}

\title{Arguments to Key Points Mapping with Prompt-based Learning}
  
\author{Ahnaf Mozib Samin \\\And
Behrooz Nikandish\\ \And
Jingyan Chen  \\ \AND
\normalfont University of Groningen \\ Groningen, The Netherlands \\
\normalfont {\texttt{\{asamin9796, behrooz.nikandish, chenjingyan0722\}@gmail.com}}
}

\date{}

\begin{document}
\maketitle

\begin{abstract}

Handling and digesting a huge amount of information in an efficient manner has been a long-term demand in modern society. Some solutions to map key points (short textual summaries capturing essential information and filtering redundancies) to a large number of arguments/opinions have been provided recently \cite{bar-haim-etal-2020a}. To complement the full picture of the argument-to-keypoint mapping task, we mainly propose two approaches in this paper. The first approach is to incorporate prompt engineering for fine-tuning the pre-trained language models (PLMs). The second approach utilizes prompt-based learning in PLMs to generate intermediary texts, which are then combined with the original argument-keypoint pairs and fed as inputs to a classifier, thereby mapping them. Furthermore, we extend the experiments to cross/in-domain to conduct an in-depth analysis. In our evaluation, we find that i) using prompt engineering in a more direct way (Approach 1) can yield promising results and improve the performance; ii) Approach 2 performs considerably worse than Approach 1 due to the negation issue of the PLM.

\end{abstract}

\section{Introduction}
With internet technology getting more accessible to the general public, a flood of information in the digital space can be observed. On online social media, people tend to provide arguments/counter-arguments on various topics, including government policies, movie reviews, and controversial issues such as gun control, abortion, and global climate changes, etc. This kind of information is valuable for government policymakers, business people, and academicians who conduct research on societal changes over time. However, due to the abundance of arguments, it becomes nearly impossible to go through each one manually and make a decision. Moreover, manually reading the arguments does not allow systematic categorization, making it unlikely to quantify them.

To address the issue, \citet{bar-haim-etal-2020a} first proposed a method to categorize the arguments by mapping them to a set of pre-defined key points set by the domain experts. They fine-tuned a pre-trained language model (PLM) using the ArgKP dataset they built. Fine-tuning PLMs has been proved to achieve superior results over the conventional approach of training a neural network model from scratch. However, there are several limitations to directly fine-tuning PLMs. First, fine-tuning a PLM requires a substantial amount of data and computational resources for each downstream task. Second, the typical way of directly fine-tuning the PLMs does not simulate how the human brain performs NLP tasks. Humans need to be prompted by providing additional task-specific information at first. For example, if we want to know whether a review is positive, negative, or neutral from a human, we would prepare a question like "Do you think the review is positive, negative, or neutral?" to prompt the human to accomplish the task. 

Prompt-based learning, built on language models that model the probability of text directly, has been a recent revival in NLP and has shown great potential to address the above limitations. \citet{brown2020language} indicated that developing a very large PLM with 175 billion tokens and prompting the PLM alleviates the need for additional data for fine-tuning. Thus, it allows us to perform zero-shot and few-shot learning for several NLP tasks. Motivated by this, we exploit prompt-based learning to accomplish the argument-to-keypoint summarization task. More precisely, we would like to shed light on the following research questions:
 \begin{itemize}
     \item Does prompt-based learning allow better utilization of the PLMs for the argument-to-keypoint mapping task? In other words, can it outperform the typical direct fine-tuning PLMs approach? 
     \item What are the challenges that arise with implementing prompt-based learning for this task?
 \end{itemize}
 
Our contributions are mainly two-fold:

\begin{itemize}
    \item First, we implement prompt-based learning for the argument-to-keypoint mapping task for the first time, to the best of our knowledge, and compare the results with conventional fine-tuning approaches. To this end, we fine-tune the T5-base PLM with five different prompt templates and report their final F1-scores.
    
    \item Second, we propose a novel architecture that takes an argument as input using prompt-based learning and generates an intermediary text after fine-tuning the PLMs. Then, we employ several machine learning classifiers to decide whether the argument, key point, and the intermediary text triple are a match. We demonstrate and analyse the promising results and shortcomings of the proposed architecture.

\end{itemize}


\section{Related Work}

Some researchers have done well-executed and rigorous studies and provided thoughtful methods in the field of argument-to-keypoint summarization. 
\citet{bar-haim-etal-2020a} established the ArgKP dataset which is the first large-scale dataset for this task and proposed a method to automatically map many arguments to a small number of given key points. They analysed and evaluated some unsupervised methods with TF-IDF and word embeddings and supervised methods like fine-tuning Bidirectional Encoder Representations from Transformers (BERT) \citep{DBLP:BERT}. This study made an excellent basis for next research in this field and is also the foundation of our project. To improve the performance of this task, \citet{kapadnis-etal-2021-team} leveraged existing state-of-the-art PLMs along with incorporating additional datasets (IBM Rank 30k and STS) and features like the topic of arguments. But the main shortcoming of these two studies is that the key points are pre-defined by expert annotators, which is an obstacle to making the process fully automatic. 

Later, \citet{bar-haim-etal-2020b} made a more in-depth study to promote the previous line of research, and developed a method for extracting key points automatically from a set of comments, which allows fully automatic key point analysis. And they compared more PLMs including  BERT-large-uncased \citep{DBLP:BERT}, XLNett-large-cased \citep{NEURIPS2019_Xlnet},  RoBERTa-large \citep{Liu2019RoBERTaAR}, and ALBERT-xxlarge-v1 \citep{Albert}, in terms of run time and accuracy, which showed a significant improvement above their previous best results in \citet{bar-haim-etal-2020a}. However, during the step of the automatic key point extraction process, they considered only single sentences and filtered out long sentences as well as those sentences that start with pronouns. Consequently, the model likely misses some potential key points.

Prompt engineering has recently become an emerging field of study in NLP. \citet{Liu2021Prompting} introduced the basics of this new paradigm in detail, and  \citet{brown2020language} confirmed the advantages of adopting prompt-based learning on various NLP tasks such as question answering, translation, and probing tasks for common sense reasoning. And prompt engineering techniques also work well on probing factual knowledge in language models \citep{jiang-etal-2020-x}. Nonetheless, the suitability of using prompt-based learning for a wide variety of NLP tasks has yet to be proven, and prompt-based learning has not been explored deeply in argument to key point summarization. In this work, we employ this promising paradigm to improve the task of argument-to-keypoint mapping further and provide two approaches to examine the performance of prompt-based learning.

\section{Methods} 

We explore two approaches for this task and compare them with our baselines without any prompt engineering technique. Approach 1 aims to make classification using prompt-based learning. And Approach 2 consists of text generation and text classification with the help of prompt engineering. We use three different Transformer-based \citep{Vaswani2017} PLMs (BERT \citep{DBLP:BERT}, BART \citep{lewis-etal-2020-bart} and T5 \citep{T5-colin}) to implement these approaches. The following subsections describe how these PLMs work and why they are appropriate for this task. Moreover, we introduce prompt engineering and the structure of the two approaches in this section.\\

\subsection{Pre-trained Language Models}
\paragraph{BERT}
Unidirectional pre-train architectures limit the choice of architectures during pre-training. For instance, utilizing left-to-right architecture like in OpenAI GPT \citep{radford2018GPT}, each token can only attend to previous tokens in the self-attention layer of the Transformer. \citet{DBLP:BERT} proposed BERT to alleviate the limitations of unidirectional architectures using \textit{a masked language model}. The architecture of the model is a multi-layer bidirectional Transformer encoder. The model is pre-trained utilizing two unsupervised tasks: Masked Language Models and Next Sentence Prediction (NSP). In many downstream tasks as well as the argument-to-keypoint task, understanding the relationship between two sentences is critical. The BERT model is pre-trained for a binarized NSP task to train the model to understand sentence relationships, which makes the BERT model a good choice for the key point analysis task. 

\paragraph{BART}
BART is a PLM that combines Bidirectional and Auto-Regressive Transformers \citep{lewis-etal-2020-bart}. The denoising autoencoder is built using a sequence-to-sequence model and it can be applied to various downstream tasks. It uses a standard Transformer-based neural machine translation architecture with a bidirectional encoder and a left-to-right decoder. During the pre-training process, an arbitrary noise function is applied to the input text, and then a sequence-to-sequence model is responsible for reconstructing the original text. Section \ref{approach2} describes Approach 2 in which our model generates an intermediary text. BART can be a reasonable choice for this task because it performs effectively in text generation \citep{BART-text-generation} and text summarization \citep{BART-text-summarization} tasks.\\

\paragraph{T5}
\citet{T5-colin} proposed a unified text-to-text Transformer-based model to explore the limitations of transfer learning using an encoder-decoder architecture. It comprises an encoder that maps the input words from the source language to an output representation. The decoder is a conditional language model that attends to the encoder representation and generates target words one by one, based on the source word and previously generated target language words at each time step. The main idea behind this model is to consider all text processing tasks as a text-to-text problem, feeding the model a text as input and generating new text as output. This provides the ability to apply the model, loss function, hyperparameters, and other parameters to various tasks, including machine translation, text summarization and classification, and question answering. We plan to use T5 in both Approach 1 and 2.

\subsection{Baseline}
\begin{figure}[t]
    \centering
    \includegraphics[scale=0.8]{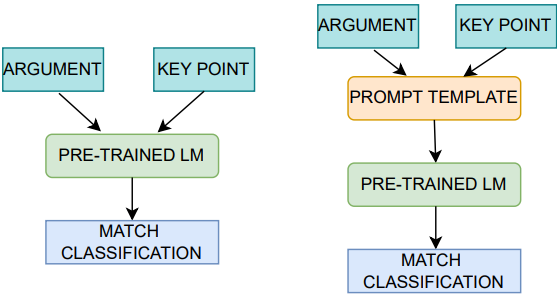}
    \caption{The architecture of baseline (left) and Approach 1 (right)}
    \label{fig:app1}
\end{figure}

The architecture of our baseline which is shown in Figure \ref{fig:app1} on the left is similar to \citet{bar-haim-etal-2020a}'s work. We build a classifier to identify whether a pair of \textit{(argument, key point)} is matched or not. To this end, we fine-tune four PLMs, BERT-base, BERT-large, T5-base, and T5-small. We train the models with the train set first and adjust the parameters like epoch with the dev set, and the trained model is evaluated using the unseen data from the test set finally. We do not employ any prompt engineering in the baselines to compare our results with other prompt-based learning approaches in this work.

\subsection{Prompt Engineering} \label{prompting}
To train a model in traditional supervised learning, it is required to have large amounts of supervised data for the task at hand. Prompt-based learning approaches are an attempt to get around this problem. We will first explain the basic form of prompting, and then show how we adopt prompting techniques in the argument-to-keypoint mapping task.\\
\citet{Liu2021Prompting} described the basic prompting process in three steps: The first step is \textit{prompt addition}, in which a \textit{prompting function} is defined to pre-process the input text. This step consists of two processes:
\begin{enumerate}
    \item Creating a \textit{template}, which consists of some fixed extra tokens and two slots: \textit{input slot} [X] for input text and \textit{answer slot} [Z] for predicted output that will be used in the \textit{answer mapping} step.
    \item Filling input slot [X] with the input text.
\end{enumerate}
The output slot [Z] could be either in the middle of the template (\textit{cloze prompt}) or at the end (\textit{prefix prompt}). Depending on the task, the number of input and output slots can vary freely. The second step is \textit{answer search}. In this step, the output slot [Z] in the prompt will be filled by a potential answer, which is the highest scoring answer. In the last step, \textit{answer mapping}, the highest-scoring answer will be mapped to the highest-scoring output. This is the case in text generation tasks, but in some tasks like text classification, each potential answer has a corresponding output to be mapped to.
\\
To answer the research question, we use prompt engineering techniques in two separate approaches for cross/in-domain to investigate if prompt-based learning can outperform our baselines. The architectures of our approaches are discussed in the following subsections.

\begin{figure}[t]
    \centering
    \includegraphics[scale=0.78]{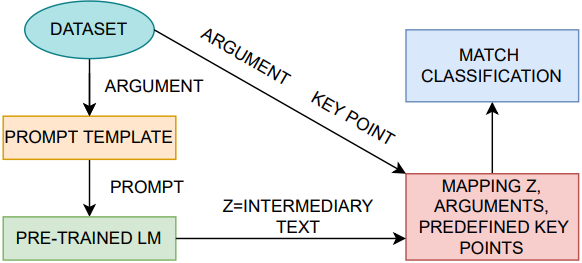}
    \caption{The architecture of Approach 2}
    \label{app2}
\end{figure}

\subsubsection{Approach 1} \label{approach1}
In Approach 1, we use prompt-based learning in conjunction with fine-tuning a PLM. As the architecture illustrated in Figure \ref{fig:app1} on the right, the \textit{(argument, key point)} pairs are transformed to the given prompt template and fed to the T5-base as input. We try various templates shown in table \ref{table_dataset}  to see how different templates influence our results. In these templates, taking \textit{(argument, key point)} as input texts, [X1] and [X2] represent the \textit{argument} and the \textit{key point} respectively. The answer space of [Z], which is the output text, can be either \textit{matched/not matched} or \textit{Yes/No}, depending on the chosen template. The following example shows the process of creating an input text using a prompt template:
\begin{itemize}
    \item \textbf{Argument [X1]}: \textit{Urbanization destroys the enviroment, and mankind should be finding ways of utilising the space already occupied more efficiently instead}
    \item \textbf{Key point [X2]}: \textit{Urbanization harms the environment}
    \item \textbf{Answer space [Z]}: \textit{matched, not matched}
    \item \textbf{Template}: \textit{The argument: [X1] is [Z] with the key point: [X2]}
    \item \textbf{Input text}: \textit{The argument: urbanization destroys the enviroment, and mankind should be finding ways of utilising the space already occupied more efficiently instead, is \textbf{matched} with the key point:  Urbanization harms the environment.}
\end{itemize}

\begin{flushleft}
\begin{table*}[h]
\begin{center}
\vspace{0.1cm}
\setlength{\tabcolsep}{2.8pt}
\begin{tabular}{ccccc}
\hline
\textbf{Topic} & \textbf{Argument} & \textbf{Key point} & \textbf{Stance} & \textbf{Label} \\
\hline
\makecell[l]{We should abandon \\the use of school uniform} & \makecell[l]{we should not abandon the use\\ of school uniforms because \\it allows children to not be\\ concerned with competitiveness \\while attempting to learn.} & \makecell[l]{Children can still \\express themselves \\using other means} & -1 & 0 \\
\hline
\makecell[l]{We should adopt atheism} & \makecell[l]{we should adopt atheism \\because religion causes too \\ much tension and disagreements.} & \makecell[l]{Atheism should be\\ adopted since we cannot \\prove that God exists} & 1&0\\
\hline
\makecell[l]{We should end mandatory\\ retirement} & \makecell[l]{mandatory retirement is a good\\ way of refreshing the workforce,\\ motivating those lower in the\\ pecking order and creating\\ employment opportunities.} & \makecell[l]{A mandatory retirement\\ age creates opportunities\\ for other workers} & -1 & 1\\
\hline
\end{tabular}
\caption{Examples from the ArgKP dataset. \textbf{Stance} means the argument support(1) or oppose(-1) the topic; \textbf{Label} represents the key point is matching (1) or non-matching (0) with the argument.}
\label{tab:data}
\end{center}
\end{table*}
\end{flushleft}

\begin{table*}[h]
\vspace{0.5cm}
\begin{center}
\setlength{\tabcolsep}{4.0pt}
\begin{tabular}{ccrr|rr|rrrr}
\hline\rule{0pt}{8pt}
& & \multicolumn{6}{c}{\textbf{Number of samples per set}} & \\
\cline{3-8}
\textbf{Experiment} & \textbf{Class} & \multicolumn{2}{c|}{\textbf{Train}} & \multicolumn{2}{c|}{\textbf{Dev}} & \multicolumn{2}{c}{\textbf{Test}} & \multicolumn{2}{c}{\textbf{Total samples}}\\[0pt]
& & \textbf{Count} & \textbf{(\%)} & \textbf{Count} & \textbf{(\%)} & \textbf{Count} & \textbf{(\%)} & \textbf{Count} & \textbf{(\%)}\\
\hline\rule{0pt}{8pt}
\multirow{3}{*}{Cross-domain} 
 & All & 17,019 & 70.6 & 2,903 & 12.1 & 4,171 & 17.3 & 24,093 & 100.0 \\
& Matching & 3,510 & 14.5 & 728 & 3.0 &  760 & 3.1 & 4,998 & 20.7 \\
& Non-matching & 13,509   & 56.1 & 2,175 & 9.1 & 3,411 & 14.2 & 19,095 & 79.3\\[2pt]
\hline
\multirow{3}{*}{In-domain} 
 & All & 17,021   & 70.6 & 2,904 & 12.1 & 4,168& 17.3 & 24,093 & 100.0 \\
& Matching &3563 &  14.8 &593& 2.5 & 842& 3.5 & 4998&20.7\\
& Non-matching& 13458 & 55.8 & 2311 & 9.6 & 3326 &13.8& 19,095 & 79.3\\[2pt]
\hline
\end{tabular}
\end{center}
\caption{Dataset distribution for cross/in-domain experiments}
\label{table_dataset}
\end{table*}

\subsubsection{Approach 2} \label{approach2}
For Approach 2, we want to explore the effect of adding additional context based on the prior knowledge of the PLMs. Figure \ref{app2} illustrates the architecture of Approach 2. First, the input argument is transformed into the given prompt template as [X] and then is fed to the trained PLM (T5-small or BART-large), which is used for text summarization. The output slot [Z] is filled by a generated summary that is called as \textit{intermediary text} in this paper. To note that we use different templates displayed in Table \ref{table_approaches} for matching/non-matching pairs to generate the corresponding intermediary texts, and the templates for two types of pairs have totally opposite connotations (e.g., \texttt{mean-not mean} and \texttt{correct-wrong}). Lastly, different classifiers are built to determine whether the generated intermediary text, argument, and keypoint triple is matching or not. In this step, we fine-tune BERT-base and T5-small PLMs and apply three machine learning algorithms (Naive Bayes \citep{mccallum1998comparison}, Support Vector Machine (SVM) \citep{cortes1995support}, Decision Tree \citep{10.1023/A:1022643204877}) with TF-IDF features. 


\section{Experiments}
This section will introduce the ArgKP dataset we use and elaborate on how we processed the data and carried out the experiments.

\subsection{Data}\label{data}

We use the established ArgKP dataset in this project \citep{bar-haim-etal-2020a}. The arguments in ArgKP revolve around 28 disputed topics, and they are a subset of the IBM-Rank-30k dataset \citep{gretz2020large}. The key points were authored by an expert on those topics. Crowd annotations were gathered to see if a keypoint represented or matched an argument, which resulted in \textit{(argument, key point)} pairs. As shown in the Table \ref{tab:data}, each pair is assigned a matching or non-matching label and a stance towards the topic.

There are 24,093 labeled argument-keypoint pairs, and 20\% of them are matching/positive pairs. Table \ref{table_dataset} displays the distribution of each data split set for cross/in-domain experiments. For the cross-domain experiments, we split the whole dataset according to the number of topics, and each topic only occurs once. We assign 19 topics to the train set, and the dev and test sets contain 4 and 5 topics, respectively. The argument-keypoint pair ratio of the three sets is 71:12:17. For the in-domain experiments, we use the same pair ratio of three split sets as the cross-domain experiments, and each split set includes all of those 28 topics.

\subsection{Pre-processing}
\raggedbottom
The ArgKP dataset is well-structured and clean enough so that we do not do much pre-processing except for some basic steps. Some arguments and all key points in the dataset do not contain full stops at the end of the sentence, so our first step is to add full stops for each full sentence if they are missing. The second step is tokenization. The PLMs (BERT, BART, T5) we mainly utilize expect a sequence of tokens as an input, so the tokenizers those PLMs were trained on are employed to tokenize the texts. For the machine learning algorithms (SVM, Naive Bayes, Decision Tree), we tokenize the texts and remove stop words using NLTK python package \citep{bird2009natural}.

\subsection{Experiment Setup}
We implement Approach 1 using OpenPrompt framework\footnote{\url{https://github.com/thunlp/OpenPrompt} }, which is an extensible and open-source toolkit for prompt engineering \citep{ding2021openprompt}. We replicate their code to train T5-base using the ArgKP dataset for this task. The code associated with this paper is available on a GitHub repository.\footnote{\url{https://github.com/samin9796/arg2keypoint}}

Table \ref{parameter} contains the some of the hyperparameters of each PLM that is fine-tuned in the baselines and Approach 1 and 2.

\begin{table}[hbtp]\centering
\setlength{\tabcolsep}{3.0pt}
\begin{tabular}{c|c|c|c}
\hline
\textbf{PLM} & \textbf{Learning Rate} & \textbf{Epoch} & \textbf{Optimizer}\\
\hline
\makecell[l]{BERT-base\\BERT-large} & 2e-5 & 3 & Adam\\
\hline
\makecell[l]{T5-base} & 1e-3/1e-4 & 3 & Adam\\
\hline
\makecell[l]{T5-small} & 3e-4 & 4 & Adam\\
\hline
\makecell[l]{BART-large} & 2e-5 & 5 & Adam \\
\hline
\end{tabular}
\caption{Hyperparameters used for finetuning different PLMs}
\label{parameter}
\end{table}

\subsection{Evaluation}
Only about 20\% pairs in the dataset are matching/positive pairs, which means the class distribution is quite imbalanced, and standard metrics such as classification accuracy would be misleading in our case. Therefore, we adopt the macro-averaged F1-score, which takes the arithmetic mean of all the per-class F1-scores as the evaluation method.

In addition, we also attempt threshold metrics in order to handle the imbalance problem. Thresholds are learned from the dev set by maximizing the macro-averaged F1-score. Pairs whose matching score exceeds the learned threshold are considered matched. However, we think it is unfair to compare the results of Approach 1/2 and the baselines with different thresholds. Furthermore, most of the learned thresholds are 0.5, which is the same as the default threshold of binary classification. Accounting for these reasons, we ignore threshold metrics finally.

\begin{table*}[htb!]
\vspace{0.1cm}
\begin{center}
\setlength{\tabcolsep}{0.5pt}
\begin{tabular}{c|c|c|c|cc}
\hline\rule{0pt}{8pt}
& \multirow{3}{*}{\textbf{Prompt Template}} & \multirow{3}{*}{\textbf{\makecell{PLM for\\ Intermediary\\ Text}}} & \multirow{3}{*}{\textbf{Model}} &  \multicolumn{2}{c}{\multirow{2}{*}{\textbf{F1-score}}}\\[0pt]
& & & & \multirow{2}{*}{\textbf{in-domain}} & \multirow{2}{*}{\textbf{cross-domain}} \\[0pt]
& & & &  &  \\[0pt]
\hline\rule{0pt}{8pt}
\multirow{4}{*}{Baseline} 
 & \multirow{4}{*}{-} & \multirow{4}{*}{-} & T5-small & 0.866 & 0.700 \\
&  &  & T5-base & 0.842 & 0.682\\
&  &  & BERT-base & 0.884 & 0.720\\
&  &  & BERT-large & 0.880 & 0.709\\[2pt]
\hline
\multirow{17}{*}{Approach 1} 
 & \makecell{T1: The argument: [X1] and \\the keypoint [X2] are [Z].} & \multirow{17}{*}{-} & \multirow{17}{*}{T5-base} & \textbf{0.914} & \textbf{0.761}\\
 \cline{2-2}
 & \makecell{T2: The argument: [X1] is [Z] \\with the keypoint: [X2]} &  &  & 0.910 & \textbf{0.761}\\
\cline{2-2}
 & \makecell{T3: Does the argument: [X1] \\comprise the fact that [X2]? [Z]} &  &  & 0.908 & 0.732\\
\cline{2-2}
 & \makecell{T4: A keypoint is a summarization\\ of the corresponding argument. \\In other words, an argument \\comprises a keypoint. Does \\the argument: [X1], comprise the \\keypoint [X2]? [Z]} &  &  & 0.913 & 0.737\\
\cline{2-2}
 & \makecell{T5: Argument: [X1] Keypoint: [X2]\\ ${"soft": "Does"}$\\ ${"soft": "the", "soft_id": 1}$\\ argument matches ${"soft_id": 1}$ \\keypoint? [Z]} &  &  & 0.911 & 0.754\\[2pt]
\hline
\multirow{20}{*}{Approach 2} 
 & \multirow{5}{*}{\makecell{T6: [X1] This means [Z1].\\$[X1]$ This does not mean [Z1]}} & \multirow{5}{*}{T5-small} & Naive Bayes & 0.493 & 0.450\\
&  &  & SVM & 0.535 & 0.480 \\
 &  &  & Decision Tree & 0.543 & 0.498\\
&  &  & T5-small & 0.845 & 0.678\\
&  &  & BERT-base & 0.856 & 0.671\\
\cline{2-6}
 & \multirow{5}{*}{\makecell{T6: [X1] This means [Z1].\\$[X1]$ This does not mean [Z1]}} & \multirow{5}{*}{BART-large} & Naive Bayes & 0.502 & 0.527\\
&  &  & SVM & 0.674 & 0.513\\
 &  &  & Decision Tree & 0.629 & 0.512\\
&  &  & T5-small & 0.830 & 0.677\\
&  &  & BERT-base & 0.892 & 0.712\\
\cline{2-6}
 & \multirow{5}{*}{\makecell{T7: The correct keypoint for\\ the argument: "[X1]" is [Z1]\\The wrong keypoint for\\ the argument: "[X1]" is [Z1]}} & \multirow{5}{*}{T5-small} & Naive Bayes & 0.485 & 0.451 \\
&  &  & SVM & 0.573 & 0.499\\
 &  &  & Decision Tree & 0.549 & 0.501\\
&  &  & T5-small & 0.835 & 0.698\\
&  &  & BERT-base & 0.900 & 0.679\\
\cline{2-6}
 & \multirow{5}{*}{\makecell{T7: The correct keypoint for\\ the argument: "[X1]" is [Z1]\\The wrong keypoint for\\ the argument: "[X1]" is [Z1]}} &  \multirow{5}{*}{BART-large} & Naive Bayes & 0.500 & 0.520\\
&  &  & SVM & 0.667 & 0.529\\
 &  &  & Decision Tree & 0.614 & 0.477\\
&  &  & T5-small & 0.810 & 0.678\\
&  &  & BERT-base & 0.896 & 0.664\\[2pt]
\hline
\end{tabular}
\end{center}
\caption{Results of baselines and Approach 1 and 2 for cross/in-domain experiments}\label{table_approaches}
\end{table*}

\section{Results \& Discussion}
\subsection{Comparison between baseline and the two approaches}
Table \ref{table_approaches} shows the comparison between our baselines and the two approaches for both in-domain and cross-domain experiments. We have four baselines that do not incorporate prompt engineering. BERT-base outperforms the rest of the four models, getting an F1-score of 88.4\% in the in-domain experiment and 72.0\% in the cross-domain experiment. For Approach 1, which utilizes prompt engineering and fine-tuning T5-base with the templates, we get higher F1-scores for each of the five prompt templates examined in this study compared to our baselines. Using the five templates, we achieve almost similar F1-scores for the in-domain experiments, while variations in the F1-scores can be observed for the cross-domain evaluation. T1 template obtains the highest F1-scores with 91.4\% for the in-domain and 76.1\% for the cross-domain experiments. T2 also achieves the second-best F1-score of 91.0\% and the equal F1-score to T1. However, T3 and T4 (template with a definition of the key point) can get F1-scores below 74\%. 

As mentioned in section \ref{approach2}, our Approach 2 explores T5-small and BART-large by fine-tuning them with two prompt templates (T6 and T7) to get the intermediary texts. Then a classifier decides whether this is a match or non-match based on the argument, intermediary text, and key point as inputs. In the case of the T6 template, with fine-tuning the BART-large for getting the intermediary texts and using BERT-base as a classifier, we achieve the highest F1-scores of 89.2\% and 71.2\% for in-domain and cross-domain experiments, respectively. But the best-performing system using the T7 template utilizes T5-small to get the intermediary texts and BERT-base and T5-small as classifiers for in-domain and cross-domain experiments, respectively. The final F1-scores from the best-performing model using the T7 template are 90.0\% and 69.8\% for in-domain and cross-domain experiments, accordingly. This experiment shows that the T6 template is more suitable for BART-large, whereas the T7 template works well with T5-small. The F1-scores using the Naive Bayes, SVM, and Decision Tree as classifiers are poor compared to T5-small and BERT-base.

Comparing the best-performing models of the baselines, Approach 1 and Approach 2, it is evident that Approach 1 outperforms the baseline for both in-domain and cross-domain datasets. Approach 1 also gets substantial improvement in F1-score getting 76.1\%, compared to Approach 2, which gets 69.8\% for the cross-domain experiment. While the difference in F1-scores between Approach 1 and 2 for the in-domain experiment is minimal, with Approach 1 getting 91.4\% and Approach 2 90.0\%. We obtain a higher F1-score using Approach 2 (90\%) compared to the baselines (88\%) on the in-domain dataset, but on the cross-domain dataset, the F1-score from Approach 2 (69.8\%) is lower than the baseline (72.0\%).

\subsection{Error Analysis of Approach 2}
Table \ref{table_approaches} shows that the overall performance of Approach 2 is poor in comparison with Approach 1 for cross/in-domain experiments. We dive into the reason hidden behind this result and make two assumptions. The first assumption is that the language models used for getting the intermediary texts suffer from negation issue. As explained in section \ref{approach2}, we use slightly different templates for matching/non-matching argument-keypoint pairs to generate their intermediary texts. The templates for non-matching pairs contain a negative connotation like \texttt{not} or \texttt{wrong}, but the problem is that the PLMs (T5-small and BART-large) cannot capture negation which is demonstrated by the generated intermediary texts being almost the same regardless of whether the template is positive or negative, and thus it results in lower F1-scores from Approach 2. To make it clear, the two following intermediary text examples are extracted from the train set. Given an argument and matching and non-matching key points corresponding to the argument, we can see that their intermediary texts are the same irrespective of using two versions of prompt templates (e.g. positive and negative).
\begin{itemize}
    \item  \textbf{Argument}: \textit{by copying something you can not get a pure copy. each copy that is made is worse then the other meaning that no one knows what can happen with cloning.}
    \item \textbf{Keypoints}: \\\textbf{matching} - \textit{Cloning is not understood enough yet}\\
    \textbf{non-matching} - \textit{Cloning is unethical/anti-religious}-
    \item \textbf{Intermediary texts for both matching/non-matchin pairs}: \textit{Cloning is unnatural}
\end{itemize}

The second assumption is a decision-making process during the evaluation time. Alluded to previously, the corresponding template is selected based on matching/non-matching labels for each pair in the train set to generate intermediary texts. However, the labels are hidden in the test set, and the specific template can not be chosen. On this account, we always use the non-negative templates (\textit{This means} and \textit{The correct key point}) to get the intermediary text during the inference time.

Even though the overall results of Approach 2 are not as good as we expect, there are still some promising aspects. If the negation issue is solved successfully, Approach 2 could alleviate the need for predefined key points since it can automatically generate texts/key points.



\section{Conclusions and Future Work}


In this work, we first build the baseline models for the argument to keypoint mapping task by finetuning PLMs without implementing prompt engineering. Then, we take advantage of prompt-based learning and utilize it while finetuning PLMs with two different approaches. From the comparison between the baselines and the two specific approaches, prompt engineering substantially improves the performance of the task. However, it still includes some challenges and limitations that need to be investigated more in the case of Approach 2. To be more specific, in Approach 1, we attempt five different prompt templates with T5-base, and all of the results are better than the baselines for both cross/in-domain experiments. While in Approach 2, the intermediary texts generated from T5-small and BART-large using two prompt templates reduce the overall performance compared to baselines.

The mapping of arguments to key points can be viewed as an intermediate step toward fully automatic argument summarization. Therefore, in future work, we plan to tackle the negation problem of PLMs in Approach 2, which would be promising for generating key points automatically. Furthermore, experimenting with other sequence-to-sequence models using prompt-based learning is another interesting future direction.

\section*{Acknowledgement}
The authors would like to thank
Dr. Khalid Al-Khatib of the University of Groningen, The Netherlands for his support and assistance.
\bibliographystyle{chicago}
\bibliography{mybib}

\end{document}